\documentclass[sn-mathphys]{sn-jnl}

\usepackage{comment}
\usepackage{epstopdf}
\usepackage{indentfirst}

\jyear{2023}%

\theoremstyle{thmstyleone}%
\theoremstyle{thmstyletwo}%

\theoremstyle{thmstylethree}%

\begin{document}

\title[Multi-Modality Representation Learning via Cross-Diffusion Attention]{MutualFormer: Multi-Modality Representation Learning via Cross-Diffusion Attention}

\author[1]{\fnm{Xixi} \sur{Wang}}

\author[1]{\fnm{Xiao} \sur{Wang}}

\author*[1]{\fnm{Bo} \sur{Jiang}}\email{jiangbo@ahu.edu.cn}

\author[1]{\fnm{Jin} \sur{Tang}}

\author[1]{\fnm{Bin} \sur{Luo}}

\affil[1]{\orgdiv{School of Computer Science and Technology, Anhui University, Hefei, 230601, China}}

\abstract{
Aggregating multi-modality data to obtain reliable data representation attracts more and more attention. 
Recent studies demonstrate that Transformer models usually work well for multi-modality tasks.
Existing Transformers generally either adopt the {Cross-Attention} (CA) mechanism or simple concatenation
to achieve the information interaction among different modalities which generally ignore the issue of modality gap.
 In this work, we re-think  Transformer and extend it to MutualFormer for multi-modality data representation.
Rather than CA in Transformer,
MutualFormer employs our new design of  Cross-Diffusion Attention (CDA) to conduct the information communication among different modalities.
Comparing with CA, the main advantages of the proposed CDA are three aspects.
First, the cross-affinities in CDA are defined based on the individual modality affinities in the \textbf{metric} space which thus can naturally avoid the issue of modality/domain gap in \textbf{feature} based CA definition.
Second, CDA provides a general scheme which can either be used for multi-modality representation or serve as the post-optimization for existing CA models.
Third, CDA is implemented efficiently.
We successfully apply the MutualFormer on different multi-modality  learning tasks (i.e., RGB-Depth SOD, RGB-NIR object ReID).
Extensive experiments  demonstrate the effectiveness of the proposed MutualFormer.
%
}

\keywords{Multi-Modality Learning, CNN, Transformer, Self-Attention, Cross-Diffusion Attention}

\maketitle

\section{Introduction}\label{sec1}
As we all know, a single modality may only work well in certain situations and express monotonous and limited information. For example, the thermal camera only perceives the temperature of the object's surface; the depth camera reflects the spatial distance information; the RGB camera captures the color and texture information, but performs poor under low-illumination and fast motion, etc.
 Therefore, how to fuse information from various modalities and achieve more reliable performance draws more and more attention in recent years.

\begin{figure}[!htbp]
\centering
\includegraphics[width=1.0\textwidth]{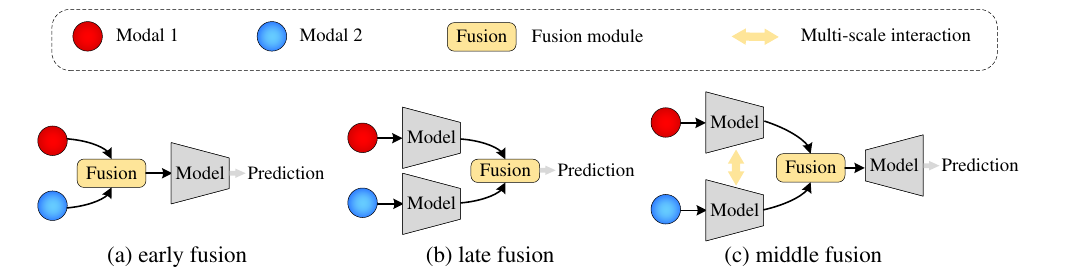}
\caption{Illustration of three fusion methods for calculating the multi-modality representation.
(a) early fusion: the different modalities are directly combined in the earliest stage for subsequent learning. (b) late fusion: the high-level features/self-predictions of different modalities are concatenated for the final prediction. (c) middle fusion: also called multi-scale fusion, which learns the interaction between different modalities by multi-scale framework or encoder-decoder architecture.
}
\label{fig:fusion_model}
\end{figure}

The fundamental challenge of multi-modality representation is \emph{how to exploit the useful cues of both intra-modality and cross-modality simultaneously for the final representation}. According to our observation, as shown in Fig.~\ref{fig:fusion_model}, existing works mainly focus on proposing various fusion methods including early fusion~\cite{ren2015exploiting,RGBD2017first}, late fusion~\cite{zhang2020select,CPFP2019CVPR,Luo2020CascadeGN}, and middle/multi-scale fusion~\cite{cmMS,piao2019depth,S2MA2020CVPR,FRDT2020MM,piao2020a2dele,chen2021cnn,wei2022multimodal}. Among them, the middle-level (or feature-level) fusion is the most popular fusion strategy due to its good balance of flexibility and accuracy.
These methods usually first utilize several separate backbone networks for various modalities to conduct intra-modality feature learning. Then, an information fusion module is designed for learning and aggregating cross-modality features. After reviewing the early deep learning algorithms on multi-modality data representation, we can find that they usually adopt CNN as the backbone, which learns the local features well with the help of convolutional operators. However, their performance is becoming saturated due to the limited representation of local CNN features~\cite{chen2021rd3d}.

Recently, Transformer has shown its strong ability in modeling the dependence of tokens (e.g., image patches) based on the token mixer architecture which is usually achieved via a self-attention module~\cite{yu2021metaformer}.
It is first proposed for natural language processing~\cite{Vaswani2017NIPS} in a self-supervised manner. Then, the Transformer based Computer Vision (CV) models~\cite{dosovitskiy2020vit,yuan2021tokenstotoken,carion2020endtoend,zhang2021bmt,touvron2021training,he2021transreid,TransT,li2022global} also obtain the top-k ranks on many benchmarks and tasks. More and more works demonstrate that Transformer can achieve comparable or even better performance than CNN. 
In addition to single modality based problems, researchers also introduce the Transformers into the multi-modality tasks \cite{tan2019lxmert, TriTransNet2021MM, liu2021visual, nagrani2021attention, MECT21ACL, dai2021Transmed}. However, existing works adopt simple concatenation \cite{TriTransNet2021MM, nagrani2021attention, dai2021Transmed}, or  cross-attention mechanism \cite{tan2019lxmert,li2020flat,liu2021visual,MECT21ACL}, which may obtain sub-optimal results. Because the first strategy generally  fails to fully exploit the relationship between the two modalities; the second way  directly connects various modalities with simple dot product, which may result in \emph{unreliable learning results caused by  the modality gap.}
Therefore, Transformer-based multi-modality  representation learning is still an interesting  problem to be further studied.

\begin{figure}[!t]
\centering
\includegraphics[width=1.0\textwidth]{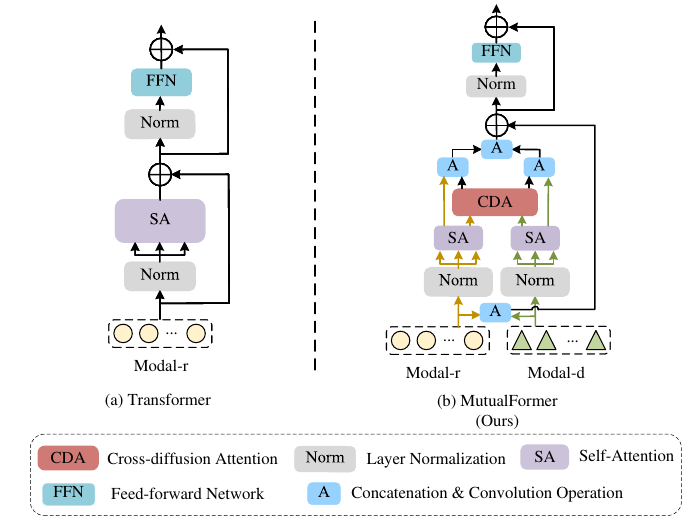}
\caption{Architectures of the proposed MutualFormer and traditional Transformer~\cite{Vaswani2017NIPS}.
As shown in (a), the traditional Transformer is generally used for single-modality (e.g., RGB image) token interaction.
Our MutualFormer mainly contains Self-Attention (SA), Cross-Diffusion Attention (CDA), FFN and Residual Connection for final token representation, as shown in (b).
The core of our MutualFormer is intra-modality token mixer with SA and cross-modality token mixer with CDA.
The details are discussed in Section~\ref{sec:mutualformer}.
Best viewed in color.
}
\label{fig:Fig.0}
\end{figure}


In this paper, we propose a novel Transformer based multi-modality data representation method, termed MutualFormer. The core of MutualFormer is our new strategy to conduct the information communication between tokens and modalities via token mixer and modality mixer.
MutualFormer contains three main modules, i.e., i) Self-Attention (SA), ii) Cross-Diffusion Attention (CDA), and iii) FFN and Residual Connection, as shown in Fig.~\ref{fig:Fig.0}.
Specifically, we first utilize SA module to capture the dependence of tokens within each intra-modality.
Meanwhile, inspired by Regularized Diffusion Process (RDP) theory~\cite{bai2018RDP}, we design a novel \emph{Cross-Diffusion Attention} (CDA) to learn the inter-modality token mixer.
Different from traditional cross-attention (CA) models which are usually defined in the \textbf{feature} space~\cite{tan2019lxmert,li2020flat,liu2021multimodal}, our CDA is defined based on individual domain affinities in the \textbf{metric} space and thus can naturally avoid the issue of {modality/domain gap in cross-modality affinities computation}.
CDA can be  used \emph{either} for multi-modality representation \emph{or} post-processing  for existing cross-attention models.
Also, CDA is usually implemented more efficiently than traditional CA~\cite{tan2019lxmert,liu2021visual}.
Finally, we adopt FFN and Residual Connection to fuse and output the context-aware representations.

Note that, the proposed MutualFormer is a general module, therefore, we can integrate or replace the  algorithm as the plug and play module. In our experiments, we take the RGB-Depth salient object detection (SOD) and RGB-NIR object re-identification tasks as two application examples to validate its effectiveness, as shown in Fig.~\ref{fig:Fig.1} and Fig.~\ref{fig:Fig.reid} respectively.
Experiments on several standard datasets demonstrate the
effectiveness of the proposed MutualFormer.

To sum up, the main contributions of this paper are summarized as follows:
\begin{itemize}
 \item We revisit the Transformer based multi-modality learning algorithms and propose a novel general MutualFormer for multi-modality data representation.
MutualFormer provides a general scheme and can be applied in many multi-modality learning tasks.

 \item We propose a novel Cross-Diffusion Attention (CDA) scheme for multi-modality information propagation.
 Compared with traditional cross-attention models, the proposed CDA is more reliable and flexible in
modeling the dependence of different modalities. 


 \item
 As the application, we integrate the MutualFormer into both RGB-D SOD and multi-modality object ReID tasks. 
Compared with many related models, the proposed MutualFormer-based architectures achieve competitive results on several public datasets. 

\end{itemize}

\section{Related Work}

\subsection{Multi-modality Transformer}
As the self-attention based Transformer network models the global long-range dependence, it can achieve comparable or better performance than CNN. Many researchers attempt to introduce the Transformer into multi-modality problems \cite{tan2019lxmert, gabeur2020multi, huang2020multimodal, dai2021Transmed, feng2021accelerated, dalmaz2021resvit, truong2021right, liu2021visual, TriTransNet2021MM, nagrani2021attention,chaudhuri2022cross,wu2023sentimental}. Part of them adopts BERT-like \cite{devlin2018bert} framework which concatenates multi-modalities along the channel dimension \cite{mao2021transformer, dai2021Transmed} or sums the tokens of each modality~\cite{TriTransNet2021MM} before input. Although good performance can be obtained, however, these methods only conduct the self-affinity computation to obtain the coarse fused features which may limit their final results.
Some works also attempt to handle this problem with cross-attention mechanism~\cite{li2020flat, MECT21ACL, zolfaghari2021crossclr}. For example, Tan $et\ al.$~\cite{tan2019lxmert} propose LXMERT framework for learning the connections of vision and language semantics for vision-and-language reasoning. Yu $et\ al.$~\cite{MT2020TIP} present a Multimodal Transformer (MT) framework by using an image encoder and a caption decoder for image captioning. Nagrani $et\ al.$~\cite{nagrani2021attention} propose Multimodal Bottleneck Transformer (MBT) used latent units to fuse the cross-modal information for audiovisual fusion. Feng $et\ al.$~\cite{feng2021accelerated} propose a multimodal transformer (MTrans) that leverages a modified attention mechanism, i.e., cross attention, to fuse the complementary information of multi-modal for accelerated magnetic resonance imaging. Curto $et\ al.$~\cite{curto2021dyadformer} develop a multi-modal multi-subject Transformer architecture (Dyadformer) to fuse both video and audio modality for dyadic interactions. Gabeur $et\ al.$~\cite{gabeur2020multi} introduce a Multi-modal Transformer (MMT) to jointly encode the representation of vision and caption in the video for video retrieval. Liu $et\ al.$~\cite{liu2021multimodal} propose a MultiModal Transformer (mmTransformer) framework based on stacked transformers to aggregate multiple channels of contextual information for multi-modal motion prediction.
Pan $et\ al.$~\cite{pan2022h} introduce a hybrid vision transformer (H-ViT) framework composed of a modal-specific controller and modal information embedding structure for multi-modality vehicle re-identification.
Wu $et\ al.$~\cite{wu2023sentimental} develop a multi-head Transformer model, termed Seti-Transformer, which leverages prior sentimental knowledge and combines both the content and sentiment information to produce the sentimental sentence.
Compared with concatenate-based multi-modality Transformers, the cross-attention based methods can model intra- and inter-modality to obtain better feature representations. Therefore, they can achieve better performance than the concatenate-based algorithms. However, the cross-modality affinities computation is generally defined by the feature representations of multi-modality token, seldom of which considers the domain/modality gap between features of different domains. Thus, these models may obtain sub-optimal results only. 

\subsection{RGB-Depth SOD}
RGB-Depth salient object detection aims to locate the most salient objects (or regions) from visual image(s). Recent SOD methods adopt Transformer or Hybrid CNN-Transformer to achieve better results, as mentioned in previous sub-sections. Before that, researchers tend to utilize CNN for feature learning and fusion.
For example, Pang $et\ al.$~\cite{HDFNet2020ECCV} propose a hierarchical dynamic filtering network (HDFNet) that mainly employs a dynamic dilated pyramid module to generate the adaptive kernel in multi-level for decoding RGB features.
Luo $et\ al.$~\cite{Luo2020CascadeGN} design Cascade Graph Neural Networks (Cas-GNN) based on Graph techniques to model the relationships between multi-modal information.
Piao $et\ al.$~\cite{piao2020a2dele} introduce a depth distiller (A2dele) to transfer depth knowledge and the localization knowledge of salient objects from the depth stream to RGB stream.
Li $et\ al.$~\cite{li2020asif} introduce an attention-steered interweave fusion network (ASIF-Net) to gradually integrates cross-modal and cross-level complementation from RGB and corresponding Depth images for RGB-D SOD task.
Ji $et\ al.$~\cite{CoNet2020ECCV} design a collaborative learning framework (CoNet) that leverages edge, depth and saliency in mutual-benefit learning manners to help generate accurate saliency results.
Liu $et\ al.$~\cite{S2MA2020CVPR} propose Selective Self-Mutual Attention to select reliable information of each modality for fusing learning.
Li $et\ al.$~\cite{HAINet2021TIP} exploit a Hierarchical Alternate Interaction Module (HAIM), which follows the RGB-depth-RGB flow for optimizing depth features based on RGB features guides. For RGB features, it adopts a hierarchical way to enhance.
Fu $et\ al.$~\cite{JLDCF2021TPAMI} propose joint learning and a densely-cooperative fusion framework based on Siamese network of sharing parameters.
Li $et\ al.$~\cite{li2022delving} design a Depth Calibration and Boundary-aware Fusion framework, which leverages a boundary-aware multimodal fusion module to fuse the information between RGB-Depth images and a learning strategy to obtain better SOD results.

\subsection{Multi-modality ReID}
The purpose of object ReID is to identify the given target image from the gallery images obtained by multiple cameras, which has attracted more and more researchers' attention. 
With the development of deep learning methods, many pioneering works~\cite{liao2017triplet,zhao2021heterogeneous,sun2018beyond,wang2018learning,zheng2022joint,shen2023git} have emerged and obtained promising performance.
But they only consider the single RGB modality, ignoring that it is highly susceptible to the illumination conditions in RGB-based ReID tasks.
Therefore, some works~\cite{barbosa2012re,munaro2014one,wu2017rgb,nguyen2017person,li2020multi,ling2020class,zheng2021robust,zheng2022multi,cao2022locality} introduce additional modality information (e.g., near-infrared/NIR, thermal infrared/TIR, depth information) to help object ReID achieve better results.
For example,
Mogelmose $et\ al.$~\cite{mogelmose2013tri} present a re-identification system based on RGB, depth, and thermal modalities for multi-modality ReID.
Wu $et\ al.$~\cite{wu2017rgb} develop a deep zero-padding network to model the interaction between RGB and NIR images for supporting person ReID task.
Wang $et\ al.$~\cite{wang2019rgb} propose an Alignment Generative Adversarial Network (AlignGAN) to exploit pixel and feature alignment jointly for the RGB-NIR ReID task.
Ren $et\ al.$~\cite{ren2019uniform} propose a uniform and variational deep learning method to fuse the visual and depth features for object ReID task.
Li $et\ al.$~\cite{li2020multi} propose a Heterogeneity-collaboration Aware Multi-stream convolutional Network to adaptively fuse different modality features in object ReID.
Ye $et\ al.$~\cite{ye2019bi} introduce a bi-directional dual-constrained top-ranking loss, which considers both the intra-modality and cross-modality variations for visible thermal object ReID task.
Ling $et\ al.$~\cite{ling2020class} propose Class-aware Modality Mix and Center-guided Metric Learning approach to smooth the discrepancy between modalities and learn discriminative cross-modality embedding for Visible thermal object ReID.
Li $et\ al.$~\cite{li2020infrared} design a X-Infrared-Visible cross-modal learning framework based on an adjoint and auxiliary X modality for infrared-visible cross-modality object ReID.
Zhai $et\ al.$~\cite{zhai2022trireid} build a TriReID dataset with text, sketch, RGB modalities and propose a Descriptive Fusion Model framework to combine the descriptive modalities of text and sketch for multi-modality ReID.
Different from them, we develop a novel MutualFormer by designing a new Cross-Diffusion Attention (CAD) to generate the multi-modality data representation for ReID task.

\section{The Proposed MutualFormer}
\label{sec:mutualformer}
Transformer has the strong ability in modeling the dependence of tokens (for example, image patches, video frames, etc) based on the token mixer architecture which is usually achieved via a self-attention (SA) module~\cite{dosovitskiy2020vit,carion2020endtoend,devlin2018bert}.
In this section, we extend this mechanism to Mutual Transformer (MutualFormer) to further model the dependence of multiple modalities for multi-modality data representation problem.
The core of MutualFormer is the design of both \emph{intra-modality token mixer} and \emph{cross-modality token mixer} to conduct the information communication between tokens and modalities respectively.
The whole architecture of MutualFormer is shown in Fig.~\ref{fig:Fig.0}, which contains 1) Self-Attention (SA) for intra-modality token mixer, 2) Cross-Diffusion Attention (CDA) for inter-modality token mixer and 3) FFN and Residual Connection for final token representation.
More details are   introduced below.

\subsection{Self-Attention (SA) for intra-modality token mixer}
\label{sec:mutualformer:SA}

Let $X_r \in \mathbb{R}^{n\times d}$ and $X_d \in \mathbb{R}^{n\times d}$ represent the token set of Modal-$r$ and Modal-$d$ respectively, where $n$ denotes the number of tokens in both modalities and $d$ is the feature dimension of tokens.
%
%
To capture the dependence of tokens within each modality, we utilize the commonly used self-attention architecture~\cite{Vaswani2017NIPS,dosovitskiy2020vit} for each modality. 
Specifically, we first compute the affinities among different tokens for Modal-$r$ as
%
\begin{align}
S_r = Softmax( \frac{Q_r K^T_r}{\sqrt{\tau}})
\label{equ:eq01} 
\end{align}
where $Q_r$ and $K_r$ are obtained by conducting two linear transformations on $X_r$ respectively and $\frac{1}{\sqrt{\tau}}$ represents scaling factor.
Based on $S_r$, we then obtain context-aware representations $U_r$ for tokens in Modal-$r$ by using the message passing  mechanism as
\begin{align}
U_r = S_rV_r 
\label{equ:eq03}
\end{align}
where $V_r$ is obtained by conducting linear transformations on $X_r$.
Similarly, we can obtain $S_d$ and $U_d$ for Modal-$d$.

\begin{figure}[!htp]
\centering
\includegraphics[width=0.68\textwidth]{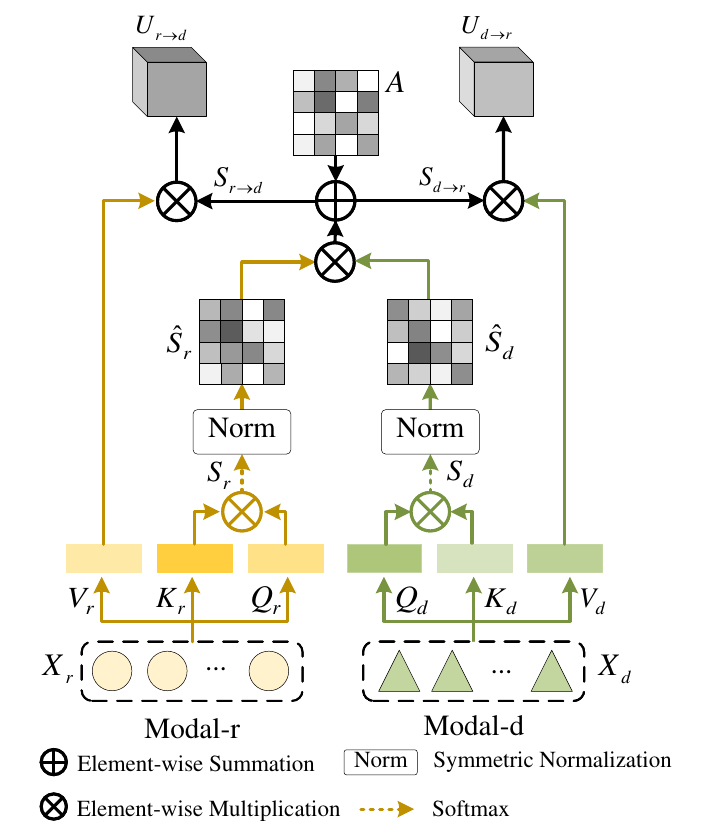}
\caption{Illustration of our proposed Cross-Diffusion Attention (CDA) module.
 $A$ is obtained by adding $S_r$ and $S_d$.
The details can be found in Section~\ref{sec:mutualformer:CDA}.
}
\label{fig:CDAModule}
\end{figure}

\subsection{Cross-Diffusion Attention (CDA) for inter-modality token mixer}
\label{sec:mutualformer:CDA}

The key aspect of the proposed MutualFormer is to conduct information communication among tokens of different modalities.
One popular way to address this issue is to design some cross-attention mechanisms~\cite{tan2019lxmert,li2020flat,liu2021visual,MECT21ACL,zhang2020context}.
The core of cross-attention is the {computation of cross-affinities} $S_{r\to d} \in \mathbb{R}^{n\times n}$ and $ S_{d \to r} \in \mathbb{R}^{n\times n}$  reflecting the dependences  among tokens of different modalities.
Previous works~\cite{tan2019lxmert,liu2021visual} generally  define cross-affinities based on features of tokens and define the Cross-Attention (CA) as,
%
\begin{align}
\label{equ:eq04} & S_{r\to d} = Softmax(\frac{Q_r {K^T_d}}{\sqrt{\tau}}) \\
\label{equ:eq05} & U_{r\to d} =  S_{r\to d}V_d
\end{align}
Similarly, we can obtain $S_{d\to r}$ and $U_{d\to r}$.
Here, we can note that the above CA involves two main steps: cross-affinities computation (Eq.(\ref{equ:eq04})) and information propagation (Eq.(\ref{equ:eq05})). The computation of cross-affinities is the core step of CA.
However,  since $Q_r$ and $K_d$  are derived from different modalities and although they are obtained by using two  linear transformations  on original input features $X_r, X_d$ respectively,
the modality/domain gap between $Q_r$ and $K_d$  can still not be ignored~\cite{cmMS,bai2018RDP}.
This intrinsic modality gap makes $(Q_rK^T_d)$ in Eq.(\ref{equ:eq04}) unreliable to reflect the pairwise affinities $S_{r \to d}$ between tokens of different modalities. 

\textbf{Cross-Diffusion Attention.}
To overcome the aforementioned issue, inspired by Regularized Diffusion Process (RDP)~\cite{bai2018RDP}, we propose to define a novel Cross-Diffusion Attention (CDA). 
{Instead of defining cross-affinities in the feature space} {in regular CA (Eq.(\ref{equ:eq04})), CDA is defined in the metric space}, as shown in Fig.~\ref{fig:CDAModule}.
%
As introduced in Section~\ref{sec:mutualformer:SA}, we can first obtain the affinities $S_r/S_d$ among Modal-$r$/Modal-$d$ tokens via self-attention in Eq.(\ref{equ:eq01}).
Then, we define $ S_{r\to d}$ based on $S_r/S_d$.
To be specific, let $\hat{S}_r = D^{-\frac{1}{2}}_rS_rD^{-\frac{1}{2}}_r$ and $\hat{S}_d = D^{-\frac{1}{2}}_dS_dD^{-\frac{1}{2}}_d$ denote the normalized affinity matrices, where $D_r, D_d$ are diagonal degree matrices. 
We propose to define our \emph{Cross-Diffusion Attention} (CDA) as
\begin{align}
\label{equ:eq06} & S_{r\to d} = \epsilon \cdot \hat{S}_rS^{(0)}_{r\to d}\hat{S}^T_d + (1-\epsilon)\cdot A \\
\label{equ:eq07} & U_{r\to d} = S_{r\to d}V_d
\end{align}
where $A = S_r+S_d$ and parameter $\epsilon \in (0, 1)$ denotes the balanced hyper-parameter.
$S^{(0)}_{r\to d}$ can be initialized as the identity matrix $I$ or some other initial affinity matrices obtained by using other approaches, such as cross-affinities of Eq.(\ref{equ:eq04}). 
In this paper, we simply set $S^{(0)}_{r\to d}$ as the identity matrix $I$. In this case, Eqs. (\ref{equ:eq06}-\ref{equ:eq07}) are simply defined as
\begin{align}
\label{equ:eq061} & S_{r\to d} = \epsilon \cdot \hat{S}_r\hat{S}^T_d + (1-\epsilon)\cdot A \\
\label{equ:eq071} & U_{r\to d} = S_{r\to d}V_d
\end{align}
%
Similarly, we can obtain $S_{d\to r}$ and $U_{d \to r}$.

 \textbf{Remark. } The main advantages of the proposed   CDA model Eqs. (\ref{equ:eq06}-\ref{equ:eq071}) are described below:
 \begin{itemize}
   \item \textbf{Reliable}. CDA is defined based on
   individual modality affinity $S_r$ and $S_d$ in metric space which thus can naturally avoid the {issue of modality/domain gap} in cross-modality affinities computation. Comparing with feature-based cross-attention (CA) methods~\cite{tan2019lxmert,li2020flat,liu2021visual}, such as Eqs. (\ref{equ:eq04}-\ref{equ:eq05}), the proposed CDA is more reliable.
   \item  \textbf{Flexible}.
   CDA provides a general and flexible learning scheme. In this paper, we simply set $A=S_r+S_d$ and $S^{(0)}_{r\to d}=I$. In real applications, one can derive many specific CDA models by initializing $A$ and $S^{(0)}_{r\to d}$ with different approaches.
   That is, CDA can also serve as a post-processing procedure for other CA methods~\cite{tan2019lxmert,li2020flat,liu2021visual} (e.g., Eqs. (\ref{equ:eq04}-\ref{equ:eq05})) to learn more reliable cross-modality attentions.
  \item \textbf{Efficient}.
  In Eq.(\ref{equ:eq061}), the affinity matrices $S_r/S_d$ are computed previously in SA module and they are shared here.
  Thus, the computation of Eq.(\ref{equ:eq061}) is faster than Eq.(\ref{equ:eq04}) because the token number $n$ is usually smaller than feature dimension $d$.
  Also, the softmax operation is not needed in  CDA.
  Therefore, our CDA is overall implemented much more efficiently than traditional CA.

 \end{itemize}

After using the above SA and CDA modules, we can obtain the representations $\{U_r,\ U_{r\to d}\}$ and $\{U_d,\ U_{d\to r}\}$ respectively for Modal-$r$ and Modal-$d$ which encode the  spatial and modality-context information  respectively. 
 By integrating them together,
 we can obtain more reliable representations $H_r, H_d \in \mathbb{R}^{n \times d}$ for two modalities as 
%
\begin{align}
\label{equ:eq08} & H_r = f_r(U_r \parallel U_{r\to d})  \\
\label{equ:eq09} &  H_d = f_d(U_d \parallel U_{d\to r})
\end{align}
%
where $\parallel$ denotes the concatenation operation on the channel dimension.
$f_r(\cdot)$ and $f_d(\cdot)$ represent  two $1 \times 1$ convolutional layers with different parameters.

\subsection{FFN and Residual Connection}

The final output $P$ of MutualFormer is 
obtained by aggregating  $H_r$ and $H_d$ together followed by feed-forword network (FFN) as
\begin{align}
& H = g(H_r \parallel H_d) + h(X_r \parallel X_d)  \\
& P = \mathrm{FFN}(\mathrm{LN} (H)) + H
\label{equ:eq10}
\end{align}
where 
$g(\cdot)$ and $h(\cdot)$ denote the two $1 \times 1$ convolutional layers and  $\parallel$ denotes the concatenation operation. 
$X_r \in \mathbb{R}^{n\times d}$ and $X_d \in \mathbb{R}^{n\times d}$ are the initial representations of the tokens, as shown in Fig.~\ref{fig:Fig.0}.
FFN denotes the two fully-connected layers with a non-linearity activation function GELU.
LN represents the layer-normalization operation.

\section{Applications}

The proposed MutualFormer provides a general learning block
for multi-modality representation learning in computer vision fields.
In this section, we apply it to two popular learning tasks, i.e., RGB-D salient object detection and RGB-NIR object Re-ID.


\subsection{RGB-Depth SOD}

\begin{figure*}[!t]
\centering
\includegraphics[width=0.9\textwidth]{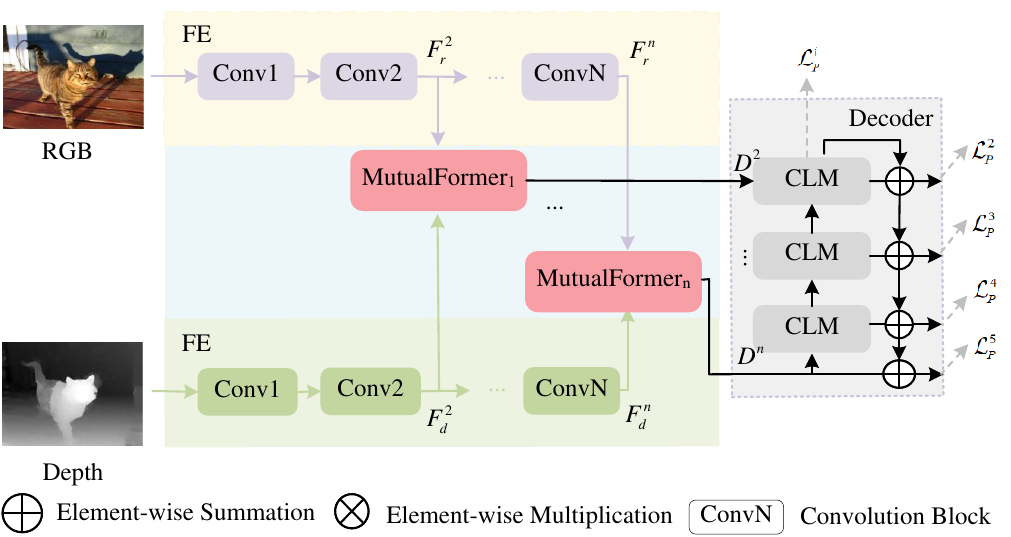}
\caption{Illustration of the RGB-Depth salient object detection based on the proposed MutualFormer.
The whole network consists of Feature Extractor (FE), MutualFormer-based representation module and Decoder.
Specifically, FE uses ResNet-50~\cite{he2016deep} pre-trained on ImageNet~\cite{deng2009imagenet} as the backbone with the supervision of focal regularization to obtain enhanced CNN representation.
$F_r^2 \backsim F_r^n$ and $F_d^2 \backsim F_d^n$ are the outputs of different levels in two branches.
The MutualFormer-based representation module refers to the proposed MutualFormer is used as the fusion module of RGB- and Depth-modality for representation learning, where the architecture of MutualFormer can be seen in Fig.~\ref{fig:Fig.0} and more details can be found in Section~\ref{sec:mutualformer}.
$D^2 \backsim D^n$ are the multi-modality feature representations of different levels after fusion.
The decoder mainly uses cross-level module (CLM) to fuse features of different levels to generate high-quality saliency maps.
Best viewed in color.}
\label{fig:Fig.1}
\end{figure*}

To validate the effectiveness of our newly proposed MutualFormer, we first select the RGB-Depth salient object detection problem.
As shown in Fig. \ref{fig:Fig.1}, overall the MutualFormer-based SOD framework contains three main modules, including Feature Extractor, MutualFormer representation module and Decoder network, as introduced respectively below.

\textbf{Feature Extractor.}  \label{subsection1}
Given the input RGB and Depth image pairs, we need to first extract their multi-level CNN feature descriptors by adopting the widely used ResNet-50 \cite{he2016deep} as the backbone network which is initialized with pre-trained models on ImageNet \cite{deng2009imagenet}.
Therefore, we can get the multi-scale feature maps $F_{r}^{l}\in \mathbb{R}^{H^{l}\times W^{l}\times C^{l}}, F_{d}^{l}\in \mathbb{R}^{H^{l}\times W^{l}\times C^{l}}, l \in \{2, 3, 4, 5 \}$ with pre-trained residual blocks.
Also, inspired by \cite{FL2020TPAMI,georgecvpr2021}, we introduce a pixel-level focal regularization term $\mathcal{L}_{R_{r\to d}}^l/\mathcal{L}_{R_{d\to r}}^l$ to alleviate the effect of clutter backgrounds or poor image quality. 

\textbf{MutualFormer-based Representation.}
We embed the aforementioned features $F^l_r, F^l_d$ into $\hat{F^l_r},\ \hat{F^l_d} \in \mathbb{R}^{ n^l \times C^l}$ via convolutional operations by following work~\cite{he2021transreid}, where $n^l$ is the total number of tokens in RGB- and Depth-modality and $C^l$ is the feature dimension of each token.
Also, we add the position encodings to its corresponding feature embedding.
After that, as illustrated in Fig.~\ref{fig:Fig.1}, we apply our newly proposed MutualFormer for generating the fused multi-scale feature representation $P^l \in \mathbb{R}^{n^l \times C^l}$ at each scale in RGB-Depth SOD, where $l=\{2,3,4,5\}$.
Finally, we use reshape and bilinear interpolation operations to restore $P^l$ to the same size as the corresponding initial feature representations at each scale, which can be denoted as $D^l \in \mathbb{R}^{H^{l}\times W^{l}\times C^{l}}, l=\{2,3,4,5\}$.

\textbf{Decoder and Loss Function. }
In the decoding phase, we use a decoder network \cite{WeiWH20} to achieve SOD prediction.
Concretely, we feed the multi-level fused features into a decoder network to obtain corresponding decoding features. The decoder contains two sub-modules.  the first sub-module takes the input features and outputs the feature map using cross-level modules (CLM)\cite{WeiWH20}. 
Then, the features predicted from the first sub-module will be used to guide the second decoding phase.
For training, we use the pixel position-aware loss~\cite{WeiWH20} for saliency prediction which consists of weighted binary cross-entropy loss and weighted intersection-over-union loss.
Formally, for each $l$-level prediction, the pixel position-aware loss $\mathcal{L}_{P}^l$ is computed as
\begin{equation}
\mathcal{L}_{P}^l = \frac{1}{2}(\omega \mathcal{L}_{ce}^{l}+\mathcal{L}_{wiou}^{l})
\label{equ:eq15}
\end{equation}
where $\mathcal{L}_{ce}^{l}$ and $\mathcal{L}_{wiou}^{l}$ denote the standard binary cross-entropy loss and weighted IOU loss~\cite{rahman2016optimizing} respectively.
$\omega = 1+5 \vert AP(y)-y \vert$ with $AP(\cdot )$ performing average pooling operation and $y$ represents ground truth.
As a result, the loss for saliency prediction is formulated as
\begin{equation}
\mathcal{L}_{P} = \frac{1}{m}\sum_{i=1}^{m}\mathcal{L}_{P}^{i}+\sum_{l=2}^{5}\frac{1}{2^{l-1}}\mathcal{L}_{P}^{l}
\label{equ:eq16}
\end{equation}
where $m$ is the total number of sub-decoder, $i$ and $l$ denote $i$-th sub-decoder and $l$-th level respectively.
By adding the regularization loss, 
the  whole loss  is
\begin{equation}
\mathcal{L}_{total} = (1-\lambda) \mathcal{L}_{P} + \lambda \sum_{l=2}^{5} (\mathcal{L}_{R_{r\to d}}^l + \mathcal{L}_{R_{d\to r}}^l)
\label{equ:eq17}
\end{equation}
where $\lambda \in [0, 0.9 ]$ stands for the balancing parameter, which is set to 0.4 in our experiments.

\subsection{RGB-NIR Object ReID}
\label{sec:reid}
\begin{figure*}[!t]
\centering
\includegraphics[width=1.0\textwidth]{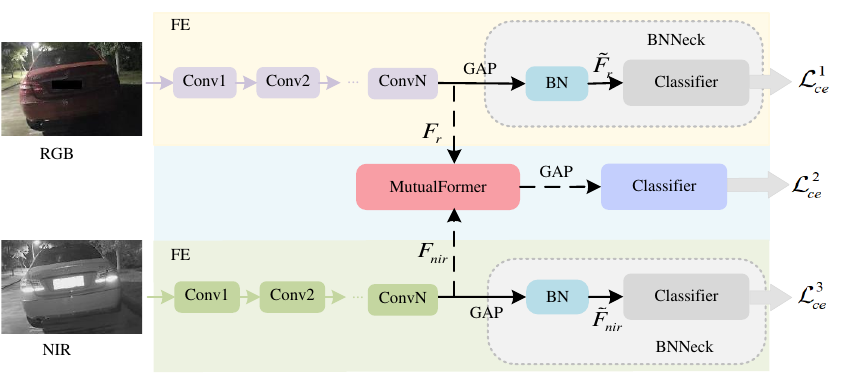}
\caption{Illustration of RGB-NIR object re-identification based on our proposed MutualFormer.
The whole network consists of Feature Extractor (FE), MutualFormer-based Fusion module and BNNeck module.
Concretely, given the input RGB and NIR modality image pairs, we first employ the ResNet-50 as feature extractor to generate the corresponding image features $F_r$ and $F_{nir}$.
Then we apply the proposed MutualFormer as the fusion module of RGB and NIR modalities to acquire the final feature representations for ID prediction as shown in the dashed lines.
Meanwhile, we add BNNeck module~\cite{luo2019bag} after the feature extractor for supervising the feature learning of different modalities.
Different colors of classifiers indicate that parameters are not shared.
}
\label{fig:Fig.reid}
\end{figure*}

To further verify the effectiveness of the proposed MutualFormer, we also employ it for the multi-modality object re-identification task. As shown in Fig.~\ref{fig:Fig.reid}, the whole framework consists of three main modules, i.e., Feature Extractor, MutualFormer-based representation module and BNNeck module. More details can be found below.

\textbf{Feature Extractor. }
We first employ the ResNet-50~\cite{he2016deep} network initialized using pre-trained weights based on the classification task on ImageNet~\cite{deng2009imagenet} dataset as the feature extractor.
Concretely, given the input RGB and NIR image pairs, we obtain their initial feature representations $F_r \in \mathbb{R}^{H \times W \times C}$ and $F_{nir} \in \mathbb{R}^{H \times W \times C}$ by using ResNet-50. $H$, $W$, and $C$ denote the height, width, and dimensionality of features for RGB and NIR modality, respectively.
Then, we introduce a BNNeck based on the global average pooling (GAP) and batch normalization (BN) layer to get the normalized RGB image features $\tilde{F}_r \in \mathbb{R}^{C}$ and the normalized NIR image features $\tilde{F}_{nir} \in \mathbb{R}^{C}$ for training by following the work~\cite{luo2019bag}.
Finally, we add the classifier with shared parameters between the RGB and NIR branch to produce the ID prediction $\hat{Y}_r=Classifer(\tilde{F}_r)$ and $\hat{Y}_{nir}=Classifer(\tilde{F}_{nir})$.


\textbf{MutualFormer-based Representation. }
In order to realize information interaction between RGB and NIR modalities in object ReID task, we apply the newly proposed MutualFormer module, as shown in Fig.~\ref{fig:Fig.reid}.
Specifically, based on the initial features $F_r, F_{nir} \in \mathbb{R}^{H\times W \times C}$ of RGB and NIR modalities mentioned above, we embed $F_r$, $F_{nir}$ into $\hat{F}_r, \hat{F}_{nir} \in \mathbb{R}^{N \times C}$ using the convolution operation by following~\cite{he2021transreid}, and add the corresponding position encodings to obtain the token set of RGB and NIR modalities, i.e., $X_r \in \mathbb{R}^{N \times C}$ and $X_{nir} \in \mathbb{R}^{N \times C}$.
$N$ and $C$ represent the total number and dimensionality of tokens for RGB and NIR modality.
Next, we leverage the proposed MutualFormer module to fuse the intra-modality and inter-modality features and thus output the fused feature representations $P$.
Finally, we add global average pooling (GAP) and the classifier for object ID prediction, which can be denoted as $\hat{Y}_{rn} = Classifer(GAP(P))$.

\textbf{Training and Testing Phase.}
Our proposed RGB-NIR object ReID framework can be optimized in an end-to-end manner.
To be specific, in the training phase, three label smoothing cross-entropy loss functions~\cite{luo2019bag} are adopted for training, including $\mathcal{L}^1_{ce}(\hat{Y}_r, Y)$, $\mathcal{L}^{2}_{ce}(\hat{Y}_{rn}, Y)$ and $\mathcal{L}^3_{ce}(\hat{Y}_{nir}, Y)$.
The loss functions can be formulated as:
\begin{equation}
\mathcal{L}_{ce} = \mathcal{L}^1_{ce}(\hat{Y}_r, Y) + \mathcal{L}^{2}_{ce}(\hat{Y}_{rn}, Y)+ \mathcal{L}^3_{ce}(\hat{Y}_{nir}, Y)
\label{equ:eq19}
\end{equation}
where $Y$ represents the ground-truth label.
The heterogeneous score coherence loss function $\mathcal{L}_{hsc}$~\cite{li2020multi} is also considered to ensure the consistency of predicted values between different modalities. Therefore, the overall loss can be formulated as:
\begin{equation}
\mathcal{L}_{total} = \mathcal{L}_{ce} + \gamma \mathcal{L}_{hsc}
\label{equ:eq20}
\end{equation}
where $\gamma$ is a hyper-parameter and is set to 0.001 in our experiments.
In the testing phase, we only need to run the MutualFormer-based fusion branch for object ReID.

%

\section{Experiments}

In this section, we demonstrate the effectiveness of MutualFormer through a large number of experiments for different multi-modality representation learning tasks, including RGB-Depth salient object detection and RGB-NIR object re-identification.

\subsection{Evaluation on RGB-D SOD}
\textbf{Datasets.}
We evaluate the proposed RGB-Depth salient detection method on six commonly used datasets, including \textbf{NJU2K}~\cite{ju2014depth} (1985 image pairs), \textbf{SIP}~\cite{fan2020rethinking} (929 image pairs), \textbf{NLPR}~\cite{peng2014rgbd} (1000 image pairs), \textbf{LFSD}~\cite{li2014saliency} (100 image pairs), \textbf{STEREO}~\cite{niu2012leveraging} (797 image pairs), \textbf{DUT-RGBD}~\cite{piao2019depth} (1200 image pairs), which are all composed of RGB and Depth image pairs.
%
Following previous works~\cite{piao2019depth, zhang2020select, piao2020a2dele}, we select 1485, 700 and 800 image pairs from NJU2K~\cite{ju2014depth}, NLPR~\cite{peng2014rgbd} and DUT-RGBD~\cite{piao2019depth} respectively as the training data.
The remaining image pairs are used as the test set.
We employ horizontal flip and random crop as used in~\cite{CPFP2019CVPR} for training data augmentation and thus obtain a final training set that contains 32835 RGB-Depth image pairs.

\textbf{Evaluation Metrics.}
We utilize five commonly used evaluation metrics, i.e., S-measure ($S_m$)~\cite{fan2017structure}, maximum F-measure ($F_{max}$)~\cite{achanta2009frequency}, maximum E-measure ($E_{max}$)~\cite{Fan2018Enhanced} and Mean Absolute Error ($MAE$)~\cite{borji2015salient, perazzi2012saliency}, to evaluate the performance of different RGB-Depth SOD models.
S-measure calculates the structure affinity between the predicted saliency result and ground truth and F-measure calculates the harmonic mean of average precision and recall values.
E-measure combines image level statistics with local pixel matching information to evaluate the saliency binary map.
MAE refers to the average pixel-wise absolute error between the saliency results and ground truth results.
The evaluation code we used is available at \url{http://dpfan.net/d3netbenchmark/}.

\begin{sidewaystable}
\sidewaystablefn%
\begin{center}
\footnotesize
\begin{minipage}{\textheight}
\caption{Quantitative comparison of our proposed model with other state-of-the-art RGB-D SOD models over six widely used datasets.
These comparison methods have publicly released codes or results.
$S_m$, $F_{max}$, $E_{max}$ and $MAE$ are evaluation metrics.
``-'' means that the method does not release test results for this dataset.
\textcolor{red}{Red} / \textcolor{blue}{Blue} indicates the $1^{st}$ / $2^{nd}$ result.}
\resizebox{\textwidth}{48mm}{
\tabcolsep=0.01cm
\begin{tabular*}{\textheight}{@{\extracolsep{\fill}}l l|c c c c c c c c c c c c c c c c c@{\extracolsep{\fill}}}
\toprule%
&
& \begin{tabular}[c]{@{}c@{}}CoNet\\ \end{tabular}  
& \begin{tabular}[c]{@{}c@{}}cmMS\\ \end{tabular}  
& \begin{tabular}[c]{@{}c@{}}D3Net\\ \end{tabular} 
& \begin{tabular}[c]{@{}c@{}}HDFNet\\ \end{tabular} 
& \begin{tabular}[c]{@{}c@{}}FRDT\\ \end{tabular} 
& \begin{tabular}[c]{@{}c@{}}CasGnn\\ \end{tabular} 
& \begin{tabular}[c]{@{}c@{}}SSF\\ \end{tabular}   
& \begin{tabular}[c]{@{}c@{}}A2dele\\ \end{tabular} 
& \begin{tabular}[c]{@{}c@{}}S2MA\\ \end{tabular}    
& \begin{tabular}[c]{@{}c@{}}HAINet\\ \end{tabular} 
& \begin{tabular}[c]{@{}c@{}}DQSD\\ \end{tabular} 
& \begin{tabular}[c]{@{}c@{}}RD3D\\  \end{tabular}  
& \begin{tabular}[c]{@{}c@{}}DCF\\ \end{tabular}   
& \begin{tabular}[c]{@{}c@{}}CDNet\\ \end{tabular}  
& \begin{tabular}[c]{@{}c@{}}JLDCF\\ \end{tabular} 
& \begin{tabular}[c]{@{}c@{}}TriTrans\\ \end{tabular} 
& {\textbf{OURS}} \\ 
\midrule
\multicolumn{2}{c|}{Paras(M)} &42.50  &-  &15.08  &44.15  &32.44  &-  &32.93  &15.02  &86.65  &59.82  &-  &46.90  &47.85  &32.93  &-  &139.55  &52.05 \\ \hline

\multirow{4}{*}{\rotatebox{90}{\begin{tabular}[c]{@{}c@{}}NJU2K\\ \end{tabular}}} 
&$S_m \uparrow$     &0.895 &0.900 &0.900 &0.908 &0.898 &0.911 &0.899 &0.871 &0.894 &0.912 &0.899 &0.916 &0.903 &0.872 &0.913 &\textcolor{blue}{0.920} &\textcolor{red}{0.922} \\
&$F_{max} \uparrow$ &0.893 &0.897 &0.900 &0.911 &0.899 &0.916 &0.896 &0.874 &0.889 &0.915 &0.900 &0.914 &0.905 &0.868 &0.915 &\textcolor{red}{0.926} &\textcolor{blue}{0.923} \\
&$E_{max} \uparrow$ &0.937 &0.936 &0.939 &0.944 &0.933 &0.948 &0.935 &0.916 &0.930 &0.944 &0.936 &0.947 &0.943 &0.915 &0.951 &\textcolor{red}{0.955} &\textcolor{blue}{0.954} \\
&$MAE \downarrow$   &0.046 &0.044 &0.046 &0.038 &0.048 &0.035 &0.043 &0.051 &0.053 &0.038 &0.050 &0.036 &0.038 &0.054 &0.039 &\textcolor{red}{0.030} &\textcolor{blue}{0.032} \\ \hline
\multirow{4}{*}{\rotatebox{90}{\begin{tabular}[c]{@{}c@{}}NLPR\\ \end{tabular}}} 
&$S_m\uparrow$     &0.908 &0.915 &0.912 &0.923 &0.914 &0.919 &0.914 &0.898 &0.915 &0.924 &0.916 &0.930 &0.922 &0.889 &\textcolor{blue}{0.931} &0.928 &\textcolor{red}{0.932} \\
&$F_{max}\uparrow$ &0.887 &0.896 &0.897 &0.917 &0.900 &0.906 &0.896 &0.882 &0.902 &0.915 &0.898 &0.919 &0.910 &0.864 &0.918 &\textcolor{blue}{0.924} &\textcolor{red}{0.925} \\
&$E_{max}\uparrow$ &0.945 &0.949 &0.953 &0.963 &0.950 &0.955 &0.953 &0.944 &0.953 &0.960 &0.952 &0.965 &0.957 &0.925 &0.965 &\textcolor{red}{0.966} &\textcolor{blue}{0.965} \\
&$MAE \downarrow$  &0.031 &0.027 &0.030 &0.023 &0.029 &0.025 &0.026 &0.029 &0.030 &0.024 &0.029 &0.022 &0.023 &0.034 &0.022 &\textcolor{red}{0.020} &\textcolor{blue}{0.021} \\ \hline
\multirow{4}{*}{\rotatebox{90}{\begin{tabular}[c]{@{}c@{}} DUT-RGBD\\ \end{tabular}}} 
&$S_m \uparrow$     &0.919 &0.912 &0.775 &0.908 &0.910 &0.920 &0.915 &0.885 &0.903 &-     &0.845 &-     &0.924 &0.880 &0.894 &\textcolor{blue}{0.934} &\textcolor{red}{0.936} \\[1.1px]
&$F_{max} \uparrow$ &0.927 &0.914 &0.742 &0.917 &0.919 &0.928 &0.924 &0.892 &0.901 &-     &0.827 &-     &\textcolor{blue}{0.932} &0.880 &0.892 &\textcolor{red}{0.946} &\textcolor{red}{0.946} \\[1.1px]
&$E_{max} \uparrow$ &0.956 &0.943 &0.834 &0.947 &0.948 &0.955 &0.951 &0.930 &0.937 &-     &0.878 &-     &\textcolor{blue}{0.957} &0.919 &0.928 &\textcolor{red}{0.966} &\textcolor{red}{0.966} \\[1px]
&$MAE \downarrow$   &0.033 &0.037 &0.097 &0.042 &0.039 &0.030 &0.033 &0.042 &0.043 &-     &0.072 &-     &0.030 &0.048 &0.049 &\textcolor{blue}{0.025} &\textcolor{red}{0.024} \\[1px]  \hline
\multirow{4}{*}{\rotatebox{90}{\begin{tabular}[c]{@{}c@{}}SIP\\ \end{tabular}}} 
&$S_m \uparrow$     &0.858 &-     &0.860 &\textcolor{blue}{0.886} &-     &-     &-     &-     &-     &0.880 &0.864 &0.885 &0.873 &0.798 &0.885 &\textcolor{blue}{0.886} &\textcolor{red}{0.894} \\
&$F_{max} \uparrow$ &0.867 &-     &0.861 &0.894 &-     &-     &-     &-     &-     &0.892 &0.865 &0.889 &0.886 &0.797 &0.893 &\textcolor{blue}{0.899} &\textcolor{red}{0.902} \\
&$E_{max} \uparrow$ &0.913 &-     &0.909 &\textcolor{blue}{0.930} &-     &-     &-     &-     &-     &0.922 &0.902 &0.924 &0.922 &0.863 &\textcolor{blue}{0.930} &\textcolor{blue}{0.930} &\textcolor{red}{0.932} \\
&$MAE \downarrow$   &0.063 &-     &0.063 &\textcolor{blue}{0.047} &-     &-     &-     &-     &-     &0.053 &0.065 &0.048 &0.052 &0.086 &0.049 &\textcolor{red}{0.043} &\textcolor{red}{0.043} \\ \hline
\multirow{4}{*}{\rotatebox{90}{\begin{tabular}[c]{@{}c@{}}STEREO\\ \end{tabular}}} 
&$S_m \uparrow$     &\textcolor{red}{0.908} &0.895 &0.899 &0.883 &0.901 &0.899 &0.893 &0.887 &0.890 &-    &-    &-     &\textcolor{red}{0.908} &0.882 &-  &\textcolor{blue}{0.905}   &\textcolor{red}{0.908} \\
&$F_{max} \uparrow$ &0.905 &0.893 &0.891 &0.880 &0.899 &0.901 &0.890 &0.886 &0.882 &-     &-    &-     &\textcolor{red}{0.909} &0.881 &- &\textcolor{red}{0.909}   &\textcolor{blue}{0.908} \\
&$E_{max} \uparrow$ &\textcolor{blue}{0.949} &0.939 &0.938 &0.930 &0.944 &0.943 &0.936 &0.935 &0.932 &-      &-    &-     &\textcolor{blue}{0.949} &0.929 &- &\textcolor{red}{0.950}     &0.947 \\
&$MAE \downarrow$   &0.040 &0.043 &0.046 &0.054 &0.043 &0.039 &0.044 &0.043 &0.051 &-     &-    &-     &\textcolor{blue}{0.037} &0.048 &- &\textcolor{red}{0.036}     &0.038 \\ \hline
\multirow{4}{*}{\rotatebox{90}{\begin{tabular}[c]{@{}c@{}}LFSD\\ \end{tabular}}} 
&$S_m \uparrow$     &0.862 &0.849 &0.825 &0.854 &0.857 &0.846 &0.859 &-     &0.837 &0.854 &0.851 &0.858 &0.856 &0.810 &0.850 &\textcolor{blue}{0.866} &\textcolor{red}{0.872} \\
&$F_{max} \uparrow$ &0.859 &\textcolor{blue}{0.870} &0.810 &0.862 &0.859 &0.847 &0.867 &-     &0.835 &0.853 &0.847 &0.854 &0.860 &0.805 &0.854 &\textcolor{blue}{0.870} &\textcolor{red}{0.879} \\
&$E_{max} \uparrow$ &\textcolor{blue}{0.907} &0.896 &0.862 &0.896 &0.903 &0.888 &0.900 &-     &0.873 &0.886 &0.878 &0.891 &0.903 &0.860 &0.887 &0.905 &\textcolor{red}{0.911} \\
&$MAE \downarrow$   &0.071 &0.073 &0.095 &0.076 &0.073 &0.073 &\textcolor{blue}{0.066} &-     &0.094 &0.079 &0.085 &0.073 &0.071 &0.089 &0.081 &\textcolor{blue}{0.066} &\textcolor{red}{0.062} \\
\botrule
\end{tabular*}}
\label{table:sod}
\end{minipage}
\end{center}
\end{sidewaystable}

\begin{figure}[!t]
\centering
\includegraphics[width=1.0\textwidth]{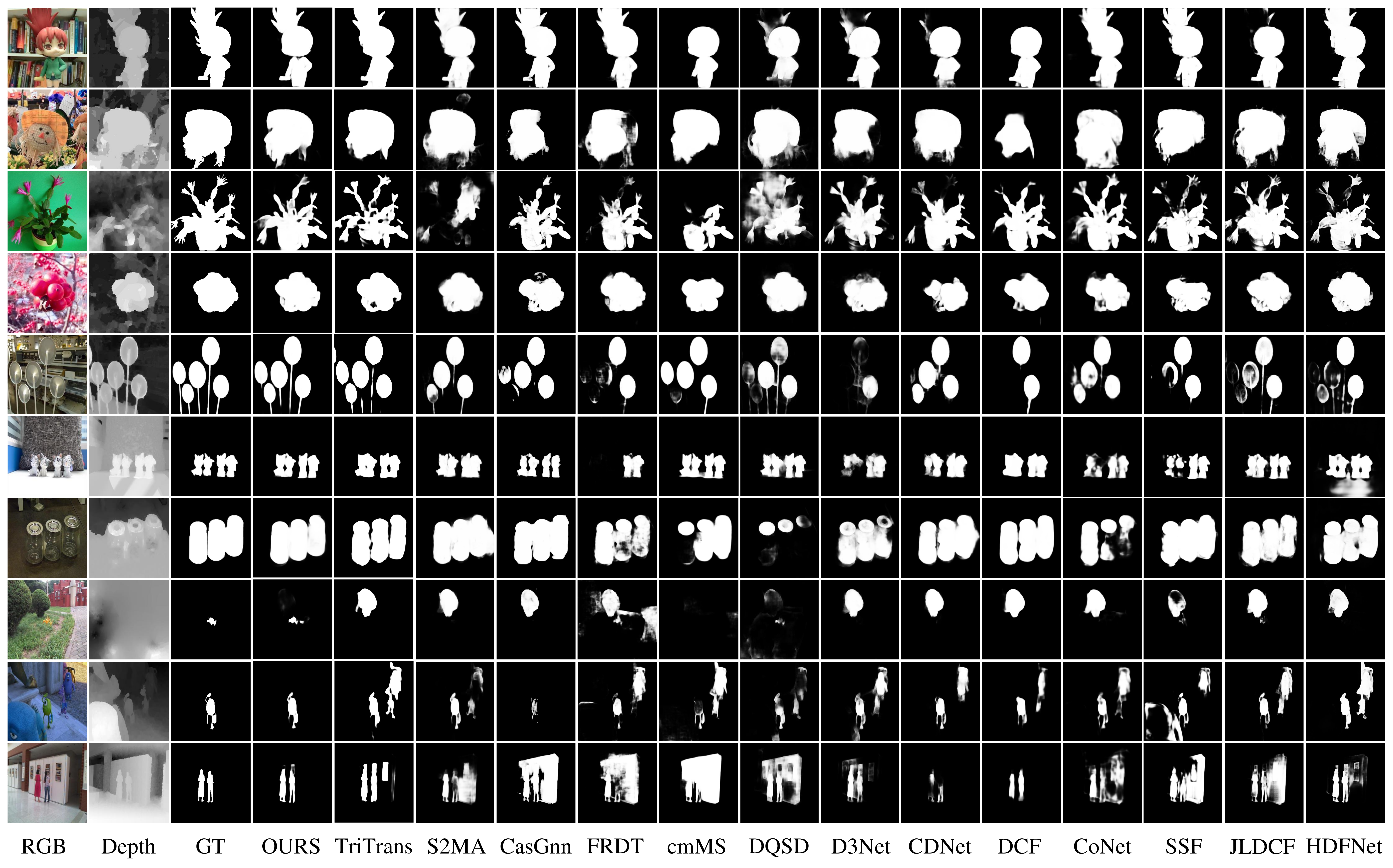}
\caption{Qualitative comparison results between the state-of-the-art RGB-Depth SOD methods and our approach.
Intuitively, one can note that the proposed MutualFormer can obtain better SOD results in various kinds of challenging scenarios. (GT: ground truth)}
\label{fig:Fig.4}
\end{figure}

\textbf{Implementation Details.}
For RGB-Depth salient object detection, we employ ResNet-50~\cite{he2016deep} with fully connected layers removed as CNN feature extractor which is pre-trained on ImageNet~\cite{deng2009imagenet} classification task.
%
All the image pairs are resized to 352 $\times$ 352 and then input to the corresponding backbone to obtain its features.
We adopt Adam~\cite{kingma2014adam} optimizer to train our model.
The initial learning rate is set to 5e-6 for the backbone and 4.5e-5 for others, which is reduced by multiplying 0.1 after 5-th epoch and 0.2 after 8-th epoch respectively.
For our fusion module, we borrow ViT~\cite{dosovitskiy2020vit} model trained on ImageNet~\cite{deng2009imagenet} to initialize the parameters of intra-modality token mixer (i.e., self-attention) and FFN in MutualFormer.
%
We experimentally set patch size as (11, 11) for all levels and the number of patches $N^l $ as $\{100, 81, 36, 1\}$ when $l = \{2, 3, 4, 5\}$.
In our model, the channel $c$ of each patch is set to 64.
%
%
%
The batch size is set to 15 and train the whole network with 15 epochs.
In the testing phase, both RGB and Depth input image pairs are first resized to 352 $\times$ 352 resolution and then feed into the proposed network to output the corresponding saliency map.
The proposed network is implemented by PyTorch 1.7 on a single 11G NVIDIA RTX2080Ti GPU.
Our code will be released at \url{https://github.com/SissiW/MutualFormer}.

\textbf{Quantitative Results.}
For fair comparisons, all the results are evaluated by using the same evaluation code provided in~\cite{fan2020rethinking}.
Table~\ref{table:sod} shows the quantitative results of four metrics on all datasets.
We can observe that our proposed network generally achieves better performance than other recent RGB-Depth SOD methods on NJU2K~\cite{ju2014depth}, NLPR~\cite{peng2014rgbd}, DUT-RGBD~\cite{piao2019depth}, SIP~\cite{fan2020rethinking}, STEREO~\cite{niu2012leveraging} and LFSD~\cite{li2014saliency} datasets. 
In particular, compared with TriTrans~\cite{TriTransNet2021MM} which is also an RGB-Depth SOD method based on Transformer.
Our proposed network achieves a better balance in both model performance and parameter number.
This demonstrates the effectiveness of the proposed RGB-D saliency detection network.
%

\textbf{Qualitative Results.}
Fig. \ref{fig:Fig.4} shows some saliency map examples of ours and the compared methods. We can find that the proposed network obtains better detection results in the various challenging scenarios, including complex backgrounds ($1^{st}$ - $2^{nd}$), salient objects and backgrounds with similar appearance ($3^{rd}$ - $4^{th}$), multiple objects ($5^{th}$ - $6^{th}$), low-quality depth map ($7^{th}$ - $8^{th}$) and small object ($9^{th}$ - $10^{th}$). It further demonstrates that our proposed method can well capture both intra-modality specific representation and cross-modality complementary information for RGB-Depth SOD task.

\subsection{Evaluation on RGB-NIR Object ReID}

\setlength{\parindent}{2em}
\textbf{Datasets. }
The RGBN300~\cite{li2020multi} dataset is used to evaluate the performance of object ReID algorithms. Many challenging factors are contained in this dataset, such as view change, occlusion, abnormal lighting, shadow, glare, low illumination, and black night. It contains 300 vehicle IDs with an average of 6.7 camera views per vehicle, and 50125 image pairs in RGB and near-infrared (NIR) modalities. The training set contains 150 vehicles, 25200 RGB-NIR image pairs. The remaining 150 vehicles, 24925 RGB-NIR image pairs, are used as the testing set. The testing set contains a gallery set and a query/probe set with 19940 and 4985 image pairs, respectively.

\textbf{Evaluation Metrics.}
To make a fair comparison, the widely used mean Average Precision (mAP) and Cumulative Matching Characteristic (CMC) are adopted as evaluation metrics.
Specifically, the mAP denotes the mean value of all queries of average precision, i.e., the region under the Precision-Recall curve.
The CMC score reflects the retrieval accuracy, which contains Rank-1, Rank-5 and Rank-10 scores in this work.

\textbf{Implementation Details.}
We use the modified HAMNet~\cite{li2020multi}\footnote[1]{Source Code: \url{https://github.com/ttaalle/multi-modal-vehicle-Re-ID}.} as our baseline model. As shown in Fig.~\ref{fig:Fig.reid}, to validate the effectiveness of the proposed MutualFormer, we utilize the MutualFormer-based representation module to replace the CAM-based adaptive fusion module in the baseline method. We resize all input images to the resolution of $128 \times 256$ for training and testing. Data augmentation strategies are adopted in the training phase, including random horizontal flip, random crop, and random erase. The Adam~\cite{kingma2014adam} optimizer is employed to optimize the whole network. The initial learning rate is 3.5e-4, then, it decayed to 3.5e-5 and 3.5e-6 after 15 epochs and 20 epochs respectively. The batch size is set to 32 and the total number of epochs is set to 30. During the testing phase, we concatenate the features output from the MutualFormer-based fusion module and the features obtained after BNNeck~\cite{luo2019bag} module as the final feature representation for evaluation. All the experiments are implemented based on PyTorch 1.7.1 and run on a server with GeForce RTX 3090Ti GPU.

\begin{table}[!t]
\centering
\caption{Comparison results on RGBN300 dataset.
`Baseline' means the result that we train the modified HAMNet~\cite{li2020multi} method with the released code.
The best results are \textbf{bolded}.}
\begin{tabular}{c|ccc|c}
\hline
\multirow{2}{*}{Methods} & \multicolumn{3}{|c|}{CMC}                                            & \multirow{2}{*}{mAP} \\ \cline{2-4}
                         & {Rank-1} & {Rank-5} & Rank-10 &                      \\ \hline
Circle Loss~\cite{sun2020circle}  & {81.0}   & {82.8}   & 84.0    & 59.3                 \\ 
PCB~\cite{sun2018beyond}          & {82.0}   & {84.7}   & 86.2    & 57.7                 \\ 
MGN~\cite{wang2018learning}       & {83.7}   & {85.5}   & 87.2    & 60.5                 \\ 
ABDNet~\cite{chen2019abd}         & {83.1}   & {-}      & -       & 58.9                 \\
HRCN~\cite{zhao2021heterogeneous} & {82.0}   & {84.7}   & 86.2    & 52.3                 \\ 
PFNet~\cite{zheng2021robust}      & {85.5}   & {87.4}   & 88.7    & 60.1                 \\ 
HAMNet~\cite{li2020multi}         & {84.0}   & {86.0}   & 87.0    & 61.9                 \\ \hline
Baseline                 & {85.9}   & {87.7}   & 89.2    & 64.9                 \\ 
\textbf{MutualFormer}             & \textbf{89.1}   & \textbf{90.6}   & \textbf{91.4}    & \textbf{69.5}                 \\ \hline
\end{tabular}
\label{table:reid}
\end{table}

\begin{figure}[!t]
\centering
\includegraphics[width=1.0\textwidth]{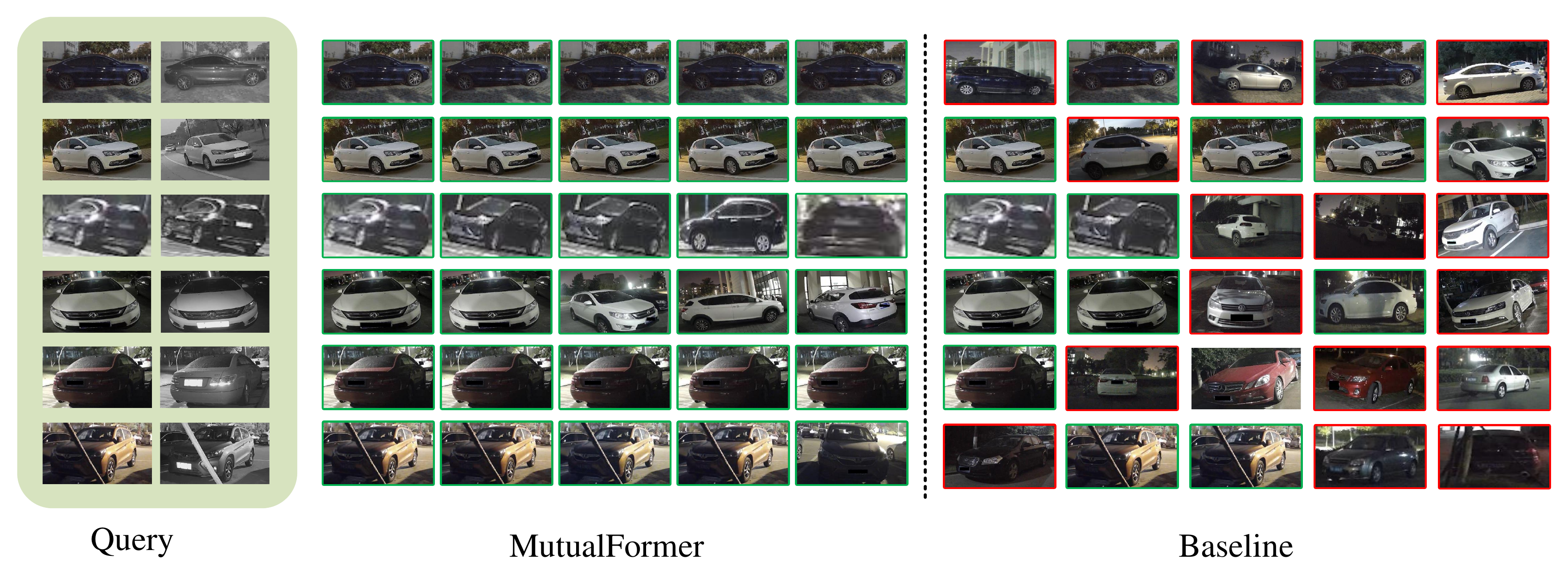}
\caption{Rank-5 results of some queries obtained by the proposed MutualFormer and Baseline model on RGBN300 dataset.
The \textcolor{green}{Green} and \textcolor{red}{Red} denote the true positive samples and false positive samples, respectively.
}
\label{fig:Fig.rank}
\end{figure}

\textbf{Quantitative Results. }
The modified HAMNet~\cite{li2020multi} is used as our baseline model and it achieves $64.9\%$, $85.9\%$, $87.7\%$ and $89.2\%$ on mAP, Rank-1, Rank-5 and Rank-10, respectively, as shown in Table~\ref{table:reid}.
In contrast, when introducing our proposed MutualFormer, the overall performance can be improved to $69.5\%$, $89.1\%$, $90.6\%$, and $91.4\%$ on mAP, Rank-1, Rank-5 and Rank-10 respectively.
It is easy to find that our proposed MutualFormer beats the CAM-based adaptive fusion module used in HAMNet~\cite{li2020multi} by $+4.6\%$ on mAP, $+3.2\%$ on Rank-1 and $+2.9\%$ on Rank-5.
When compared with other state-of-the-art models, our results are also better than theirs, for example, we outperform the PFNet~\cite{zheng2021robust} by $+9.4\%$ and $+3.6\%$ improvements on mAP and Rank-1.
We also beat the graph-based HRCN method by a large margin, i.e., $+17.2\%$ on mAP and $+7.1\%$ on Rank-1.
These results fully verified the effectiveness of the proposed MutualFormer.
Therefore, compared with traditional multi-modality fusion models, the proposed MutualFormer based on intra-modality token mixer (i.e., self-attention) and cross-modality token mixer (i.e., cross-diffusion attention) is more reliable in modeling the dependence and information communication of different modalities.



\textbf{Qualitative Results.}
To further verify the effectiveness of the proposed MutualFormer for object ReID task, as shown in Fig.~\ref{fig:Fig.rank}, we show the top-5 ranking list of ReID results on RGBN300 dataset~\cite{li2020multi}. We can see that by using the proposed MutualFormer fusion module, the ReID model can find more true positive samples than the Baseline model\footnote[1]{As shown in~\cite{li2020multi}, Baseline model adopts the CAM-based adaptive fusion module to fuse the multi-modality features.}.


\begin{sidewaystable}
\sidewaystablefn%
\begin{center}
\begin{minipage}{\textheight}
\small
\caption{Ablation study for MutualFormer Fusion layers in NJU2K, NLPR, DUT-RGBD and LFSD datasets. The best results are \textbf{bolded}.}
\resizebox{\textwidth}{14.8mm}{
\tabcolsep=0.03cm
\begin{tabular*}{\textheight}{@{\extracolsep{\fill}}c|c|c c c c|c c c c|c c c c|c c c c@{\extracolsep{\fill}}}
\toprule%
 \multirow{2}{*}{\#} &\multirow{2}{*}{} & \multicolumn{4}{c|}{NJU2K} & \multicolumn{4}{c|}{NLPR}  & \multicolumn{4}{c|}{DUT-RGBD}   & \multicolumn{4}{c}{LFSD}\\ 
               &   & $S_m \uparrow$ &$F_{max}\uparrow$ &$E_{max}\uparrow$  & $MAE \downarrow$  & $S_m \uparrow$ &$F_{max}\uparrow$ &$E_{max}\uparrow$  & $MAE \downarrow$  & $S_m \uparrow$ &$F_{max}\uparrow$ &$E_{max}\uparrow$ & $MAE \downarrow$  & $S_m \uparrow$ &$F_{max}\uparrow$ &$E_{max}\uparrow$  & $MAE \downarrow$   \\
\midrule
0 &T = 1        & \textbf{0.923} &\textbf{0.925} & \textbf{0.956}  & \textbf{0.032} & 0.928 &0.919 &0.962 & 0.022  & \textbf{0.936} &0.945 &0.965 & 0.026  & 0.865 &0.874 &0.908 & 0.064  \\
1 &T = 2        & 0.922 &0.923 & 0.954  & \textbf{0.032}  & \textbf{0.932} &\textbf{0.925} &0.965  & \textbf{0.021}  & \textbf{0.936} &\textbf{0.946} &\textbf{0.966} & \textbf{0.024}  & \textbf{0.872} &\textbf{0.879} &\textbf{0.911} & \textbf{0.062}   \\
2 &T = 3        & 0.921 &0.923 &0.953 & \textbf{0.032}  & 0.929 &0.920 &0.964 & 0.022  & 0.933 &0.941 &0.962 &  0.027  & 0.866 &0.864 &0.899 &  0.063 \\
3 &T = 4        & 0.920 &0.921 &0.950 & 0.034  & 0.930 &0.919 &0.964 & 0.022  & \textbf{0.936} &0.945 &0.965 &  0.025  & 0.866 &0.869 &0.906 &  0.064 \\
4 &T = 5        & 0.921 &0.922 &0.954 & 0.033  & 0.931 &0.922 &\textbf{0.967} & \textbf{0.021} & 0.935 &0.943 &0.962 &  0.026  & 0.867 &0.875 &0.910 &  0.064 \\
\botrule
\end{tabular*}}
\label{table:tab3}
\\
\caption{Ablation study for different fusion strategies in four datasets.
The``Add/Cat" denotes directly using a simple adding/concatenating operation to fuse the multi-modality features.
The ``CrossFormer" represents the self-attention mechanism of the standard Transformer replaced by cross-attention mechanism of Eqs. (\ref{equ:eq04}-\ref{equ:eq05}) for multi-modality fusion.
The best results are \textbf{bolded}.
}
\resizebox{\textwidth}{16.3mm}{
\tabcolsep=0.03cm
\begin{tabular*}{\textheight}{@{\extracolsep{\fill}}c|cccc|cccc|cccc|cccc@{\extracolsep{\fill}}}
\toprule%
\multirow{2}{*}{Fusion Strategy} & \multicolumn{4}{c|}{NJU2K}            & \multicolumn{4}{c|}{NLPR}             & \multicolumn{4}{c|}{DUT-RGBD}         & \multicolumn{4}{c}{LFSD}              \\
                                 & $S_m\uparrow$ & $F_{max}\uparrow$ & $E_{max}\uparrow$ & $MAE\downarrow$ & $S_m\uparrow$ & $F_{max}\uparrow$ & $E_{max}\uparrow$ & $MAE\downarrow$ & $S_m\uparrow$ & $F_{max}\uparrow$ & $E_{max}\uparrow$ & $MAE\downarrow$ & $S_m\uparrow$ & $F_{max}\uparrow$ & $E_{max}\uparrow$ & $MAE\downarrow$   \\
\midrule
Add                  &0.910 &0.907 &0.940 &0.038 &0.922 &0.908 &0.954 &0.025 &0.928 &0.932 &0.956 &0.029 &0.855 &0.863 &0.900 &0.071  \\
Cat                  &0.908 &0.902 &0.938 &0.039 &0.924 &0.912 &0.956 &0.025 &0.923 &0.922 &0.950 &0.030 &0.858 &0.854 &0.898 &0.070  \\
Transformer          &0.917 &0.917 &0.949 &0.034 &0.930 &0.923 &\textbf{0.965} &\textbf{0.021} &0.933 &0.940 &0.961 &0.027 &0.867 &0.873 &0.909 &\textbf{0.062}  \\
CrossFormer          &0.918 &0.919 &0.952 &0.034 &0.928 &0.917 &0.961 &0.023 &0.933 &0.943 &0.962 &0.026 &0.864 &0.869 &0.901 &0.064 \\\hline
CrossFormer$_{CDA}$  &0.918 &0.920 &0.952 &0.033 &0.930 &0.923 &0.964 &0.022 &0.934 &0.943 &0.965 &0.026 &0.866 &0.871 &0.904 &0.063  \\
MutualFormer         & \textbf{0.922} &\textbf{0.923} &\textbf{0.954} & \textbf{0.032}  & \textbf{0.932} &\textbf{0.925} &\textbf{0.965}  & \textbf{0.021}  & \textbf{0.936} &\textbf{0.946} &\textbf{0.966} & \textbf{0.024}  & \textbf{0.872} &\textbf{0.879} &\textbf{0.911} & \textbf{0.062} \\
\botrule
\end{tabular*}}
\label{table:tab8}
\\
\end{minipage}
\end{center}
\end{sidewaystable}

\subsection{Ablation Study}
\label{sec:ablation}
In this section, we conduct extensive experiments on four RGB-D SOD datasets (NJU2K, NLPR, DUT-RGBD and LFSD) to help readers better understand our proposed MutualFormer.

\textbf{Impact of Multi-layer MutualFormer.}
To verify the effect of multi-layer MutualFormer in each level for RGB-D SOD task, we set different layer numbers $T$ to conduct ablation study.
%
%
As shown in Table~\ref{table:tab3}, we list the comparison results of $T$ gradually increasing from 1 to 5.
We can see that the performance of $T = 2$ generally outperforms $T = 1$ except for NJU2K, which demonstrates the effectiveness of the multi-layer MutualFormer architecture at each level.
However, the performance decreases when the layer $T$ of MutualFormer is larger than 2.
As a result, we set the number of MutualFormer layers to 2 for each level in our experiments.

\textbf{Analysis of Different Fusion Strategies.}
In order to analyze the effectiveness of different fusion strategies (i.e., Add, Concatenate (Cat), Transformer, CrossFormer, CrossFormer$_{CDA}$ and MutualFormer), the entire network architecture remains unchanged except for the fusion module.
The Add, Cat, Transformer, and CrossFormer are the four commonly used multi-modality fusion strategies, where Add/Cat means directly using a simple adding/concatenating operation to fuse the two modality features.
The Transformer represents we employ the standard Transformer \cite{Vaswani2017NIPS} as the fusion module.
The CrossFormer denotes the self-attention mechanism of the standard Transformer replaced by the cross-attention mechanism (CA) of Eqs. (\ref{equ:eq04}, \ref{equ:eq05}) for multi-modality fusion.
The CrossFormer$_{CDA}$ employs our cross-diffusion attention (CDA) in Eqs. (\ref{equ:eq06}, \ref{equ:eq07}).
All the results are shown in Table~\ref{table:tab8}.
We can observe that:
1) All the transformer-based fusion methods obtain surprising results, which indicates the effectiveness of transformer architecture for the fusion multi-modality data.
2) Comparing CrossFormer$_{CDA}$ with CrossFormer, we observe that CrossFormer$_{CDA}$ can obtain better performance. For example, the CrossFormer$_{CDA}$ obtains 0.923 on $F_{max}$, meanwhile, the CrossFormer achieves 0.917 on this metric based on the NLPR dataset.
This proves the effectiveness of our proposed CDA.
3) The performance of the proposed MutualFormer is better than other methods based on transformer architecture.
This demonstrates the effectiveness of the proposed MutualFormer architecture.

\begin{sidewaystable}
\sidewaystablefn%
\begin{center}
\begin{minipage}{\textheight}
\small
\caption{Comparison results for different parameter $\epsilon$ in Eqs. (\ref{equ:eq06}-\ref{equ:eq071}).
The best results are \textbf{bolded}.}
\resizebox{\textwidth}{21.5mm}{
\tabcolsep=0.03cm
\begin{tabular*}{\textheight}{@{\extracolsep{\fill}}c|cccc|cccc|cccc|cccc@{\extracolsep{\fill}}}
\toprule%
\multirow{2}{*}{$\epsilon$} & \multicolumn{4}{c|}{NUJ2K}      & \multicolumn{4}{c|}{NLPR}      & \multicolumn{4}{c|}{DUT-RGBD}     & \multicolumn{4}{c}{LFSD}          \\
   & S$_m$ & F$_{max}$ & E$_{max}$ & MAE   & S$_m$ & F$_{max}$ & E$_{max}$ & MAE   & S$_m$ & F$_{max}$ & E$_{max}$ & MAE   & S$_m$ & F$_{max}$ & E$_{max}$ & MAE   \\ \hline
0.0             & 0.915 & 0.915   & 0.949   & 0.035 & 0.927 & 0.918   & 0.961   & 0.022 & 0.934 & 0.943   & 0.963   & 0.027 & 0.862 & 0.873   & 0.901   & 0.067 \\
0.1             & 0.918 & 0.920   & 0.950   & 0.033 & 0.926 & 0.914   & 0.959   & 0.023 & 0.934 & 0.944   & 0.963   & 0.026 & 0.863 & 0.867   & 0.900   & 0.067 \\
0.2             & 0.917 & 0.919   & 0.949   & 0.035 & 0.928 & 0.919   & 0.963   & 0.022 & \textbf{0.936} & 0.945   & \textbf{0.966}   & 0.025 & 0.868 & 0.875   & 0.907   & 0.065 \\
0.3             & 0.917 & 0.914   & 0.947   & 0.034 & 0.925 & 0.914   & 0.957   & 0.024 & 0.934 & 0.941   & 0.963   & 0.026 & 0.865 & 0.874   & 0.908   & 0.064 \\
0.4             & 0.915 & 0.913   & 0.946   & 0.035 & 0.926 & 0.917   & 0.962   & 0.022 & 0.933 & 0.941   & 0.963   & 0.026 & 0.868 & 0.870   & 0.906   & \textbf{0.062} \\
0.5             & 0.917 & 0.920   & 0.949   & 0.034 & 0.928 & 0.915   & 0.961   & 0.022 & 0.933 & 0.942   & 0.962   & 0.026 & 0.867 & 0.874   & 0.909   & 0.064 \\
0.6             & \textbf{0.922} & \textbf{0.923}   & \textbf{0.954}   & \textbf{0.032} & \textbf{0.932} & \textbf{0.925}   & \textbf{0.965}   & \textbf{0.021} & \textbf{0.936} & \textbf{0.946}   & \textbf{0.966}   & \textbf{0.024} & \textbf{0.872} & \textbf{0.879}   & \textbf{0.911}   & \textbf{0.062} \\
0.7             & 0.915 & 0.914   & 0.947   & 0.035 & 0.930 & 0.919   & 0.963   & 0.022 & \textbf{0.936} & 0.944   & 0.964   & 0.025 & 0.861 & 0.866   & 0.902   & 0.067 \\
0.8             & 0.920 & 0.922   & 0.952   & 0.033 & 0.926 & 0.917   & 0.961   & 0.023 & 0.931 & 0.941   & 0.961   & 0.028 & 0.871 & 0.874   & 0.908   & \textbf{0.062} \\
0.9             & 0.917 & 0.918   & 0.950   & 0.034 & 0.928 & 0.916   & 0.960   & 0.023 & \textbf{0.936} & 0.943   & 0.963   & 0.026 & 0.866 & 0.876   & 0.908   & 0.063 \\
1.0             & 0.918 & 0.920   & 0.950   & 0.034 & 0.928 & 0.919   & 0.962   & 0.023 & 0.934 & 0.944   & 0.963   & 0.027 & 0.864 & 0.876   & 0.906   & 0.064 \\
\botrule
\end{tabular*}}
\label{table:tab6}
\\
\caption{Comparison results of different parameter $\lambda$ in Eq. (\ref{equ:eq17}).
The best results are \textbf{bolded}.
}
\resizebox{\textwidth}{21.2mm}{
\tabcolsep=0.03cm
\begin{tabular*}{\textheight}{@{\extracolsep{\fill}}c|cccc|cccc|cccc|cccc@{\extracolsep{\fill}}}
\toprule%
\multirow{2}{*}{$\lambda$} & \multicolumn{4}{c|}{NUJ2K}        & \multicolumn{4}{c|}{NLPR}         & \multicolumn{4}{c|}{DUT-RGBD}     & \multicolumn{4}{c}{LFSD}          \\
                            & S$_m$ & F$_{max}$ & E$_{max}$ & MAE   & S$_m$ & F$_{max}$ & E$_{max}$ & MAE   & S$_m$ & F$_{max}$ & E$_{max}$ & MAE   & S$_m$ & F$_{max}$ & E$_{max}$ & MAE   \\ \hline
0.0             & 0.920 & 0.921   & 0.952   & 0.033 & 0.931 & 0.922   & 0.964   & \textbf{0.021} & 0.934 & 0.943   & 0.965   & 0.026 & 0.868 & 0.871   & 0.904   & \textbf{0.062} \\
0.1             & 0.918 & 0.920   & 0.950   & 0.033 & 0.930 & 0.919   & 0.962   & 0.022 & 0.933 & 0.941   & 0.964   & 0.026 & 0.868 & 0.866   & 0.905   & 0.065 \\
0.2             & 0.917 & 0.919   & 0.950   & 0.034 & 0.929 & 0.919   & 0.964   & \textbf{0.021} & 0.934 & 0.945   & 0.965   & 0.026 & 0.870 & 0.873   & 0.907   & 0.063 \\
0.3             & 0.917 & 0.917   & 0.951   & 0.034 & 0.929 & 0.918   & 0.962   & 0.022 & 0.935 & 0.943   & 0.964   & 0.025 & 0.870 & 0.873   & 0.910   & 0.063 \\
0.4             & \textbf{0.922} & \textbf{0.923}   & \textbf{0.954}   & \textbf{0.032} & \textbf{0.932} & \textbf{0.925}   & \textbf{0.965}   & \textbf{0.021} & \textbf{0.936} & \textbf{0.946 }  & \textbf{0.966}   & \textbf{0.024} & \textbf{0.872} & \textbf{0.879}   & \textbf{0.911}   & \textbf{0.062} \\
0.5             & 0.917 & 0.918   & 0.949   & 0.034 & \textbf{0.932} & 0.923   & 0.963   & \textbf{0.021} & 0.934 & 0.944   & 0.965   & 0.026 & 0.871 & 0.875   & 0.908   & 0.063 \\
0.6             & 0.919 & 0.921   & 0.950   & 0.034 & 0.929 & 0.921   & 0.963   & \textbf{0.021} & 0.935 & 0.944   & 0.964   & 0.026 & 0.868 & 0.874   & 0.909   & \textbf{0.062} \\
0.7             & 0.920 & 0.921   & 0.950   & 0.033 & 0.927 & 0.917   & 0.961   & 0.023 & 0.933 & 0.942   & 0.963   & 0.027 & 0.867 & 0.872   & 0.909   & \textbf{0.062} \\
0.8             & 0.917 & 0.918   & 0.947   & 0.035 & 0.928 & 0.916   & 0.960   & 0.022 & 0.935 & 0.944   & 0.964   & 0.027 & 0.867 & 0.868   & 0.905   & 0.064 \\
0.9             & 0.918 & 0.920   & 0.950   & 0.035 & 0.926 & 0.912   & 0.957   & 0.023 & 0.934 & 0.943   & 0.961   & 0.027 & 0.868 & 0.875   & 0.903   & 0.064 \\
\botrule
\end{tabular*}}
\label{table:tab5}
\\
\end{minipage}
\end{center}
\end{sidewaystable}

\subsection{Parameter Analysis}
\begin{figure*}[!t]
\centering
\includegraphics[width=1.0\textwidth]{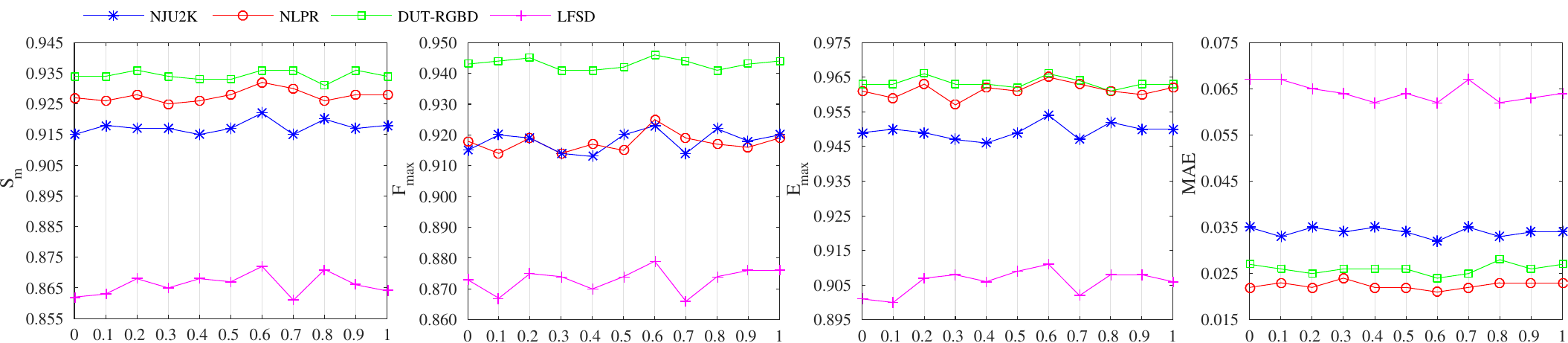}
\caption{Parameter analysis of $\epsilon$ on four datasets, including NJU2K, NLPR, DUT-RGBD, LFSD.
We set it to 0.6 in our experiment.}
\label{fig:paramepsilon}
\end{figure*}

\begin{figure*}[!ht]
\centering
\includegraphics[width=1.0\textwidth]{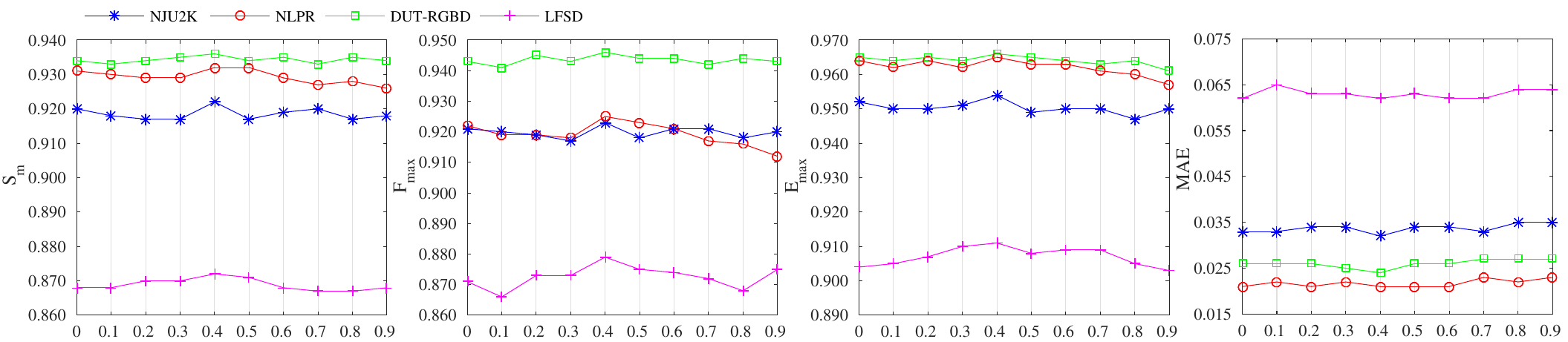}
\caption{Parameter analysis of $\lambda$ on four datasets, including NJU2K, NLPR, DUT-RGBD, LFSD.
We set it to 0.4 in our experiment.}
\label{fig:paramlambda}
\end{figure*}

In the MutualFormer-based RGB-D SOD framework, two parameters are important for the final performance, i.e., the $\epsilon$ in Eqs. (\ref{equ:eq06}-\ref{equ:eq071}), and the $\lambda$ in Eq. (\ref{equ:eq17}).
In this section, we test the MutualFormer-based RGB-D SOD model with various values of the two parameters to check their influence.
Specifically, we set the $\epsilon \in [0, 1]$ and $\lambda \in [ 0, 0.9 ]$. 
As shown in Table \ref{table:tab6} and Table \ref{table:tab5}, we can find that when the $\epsilon$ and $\lambda$ are set as 0.6 and 0.4, the MutualFormer-based RGB-Depth SOD model achieves better results on the four datasets, including NJU2K, NLPR, DUT-RGBD, and LFSD.
To give a more intuitive presentation, we also visualize these results in Fig. \ref{fig:paramepsilon} and Fig. \ref{fig:paramlambda}.

\begin{figure*}[!t]
\centering
\includegraphics[width=1.0\textwidth]{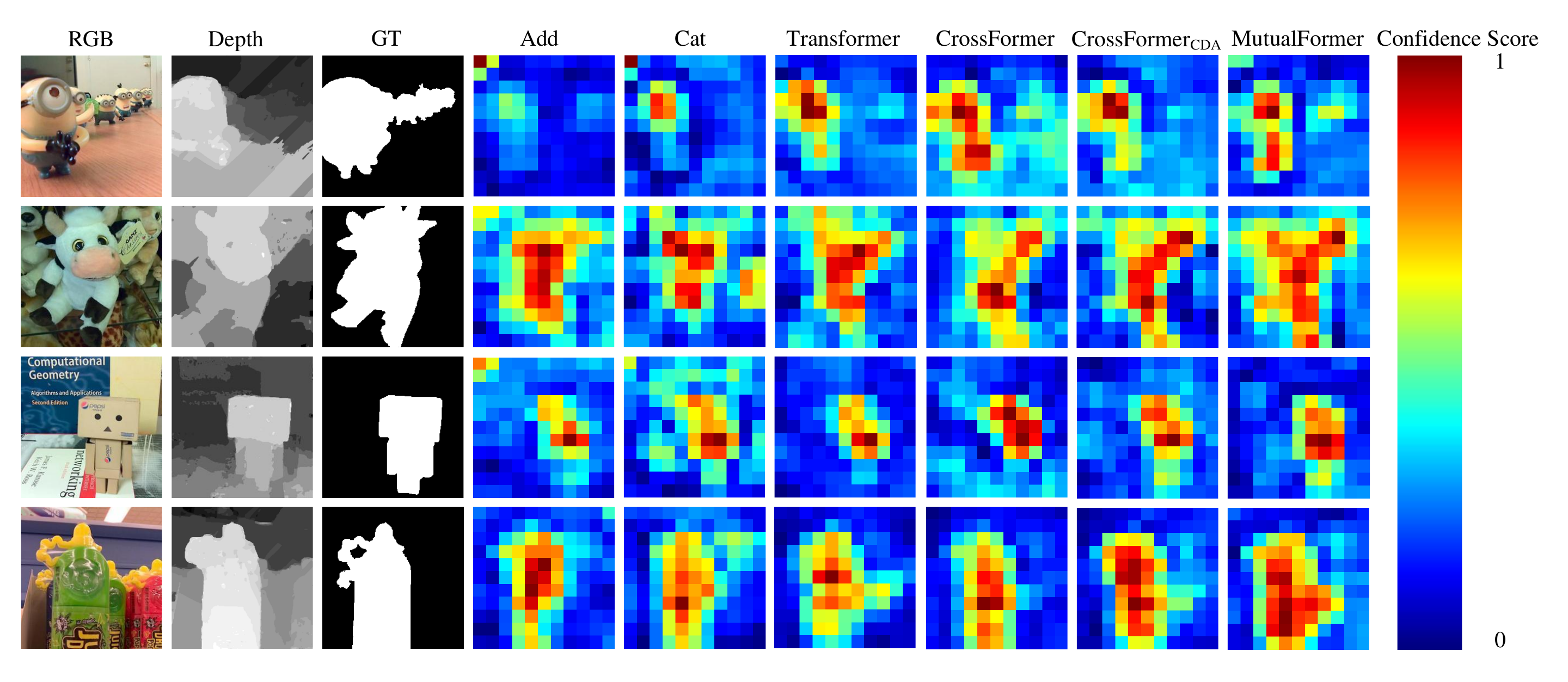}
\caption{Visualization of feature maps predicted by different fusing algorithms.
`Add/Cat' means directly using a simple adding/concatenating operation to fuse the two modality features.
`Transformer' represents we employ the standard Transformer as the fusion module.
`CrossFormer' denotes the self-attention mechanism (Eqs. (\ref{equ:eq01}-\ref{equ:eq03})) of the standard Transformer replaced by the cross-attention mechanism (Eqs. (\ref{equ:eq04}-\ref{equ:eq05})) for multi-modality fusion, while `CrossFormer$_{CDA}$' refers to employ our cross-diffusion attention (CDA, Eqs. (\ref{equ:eq061}-\ref{equ:eq071})).
We can observe that our proposed MutualFormer activates more effective object areas and thus generates better object feature map.}
\label{fig:feature_map}
\end{figure*}

\begin{figure*}[!t]
\centering
\includegraphics[width=1.0\textwidth]{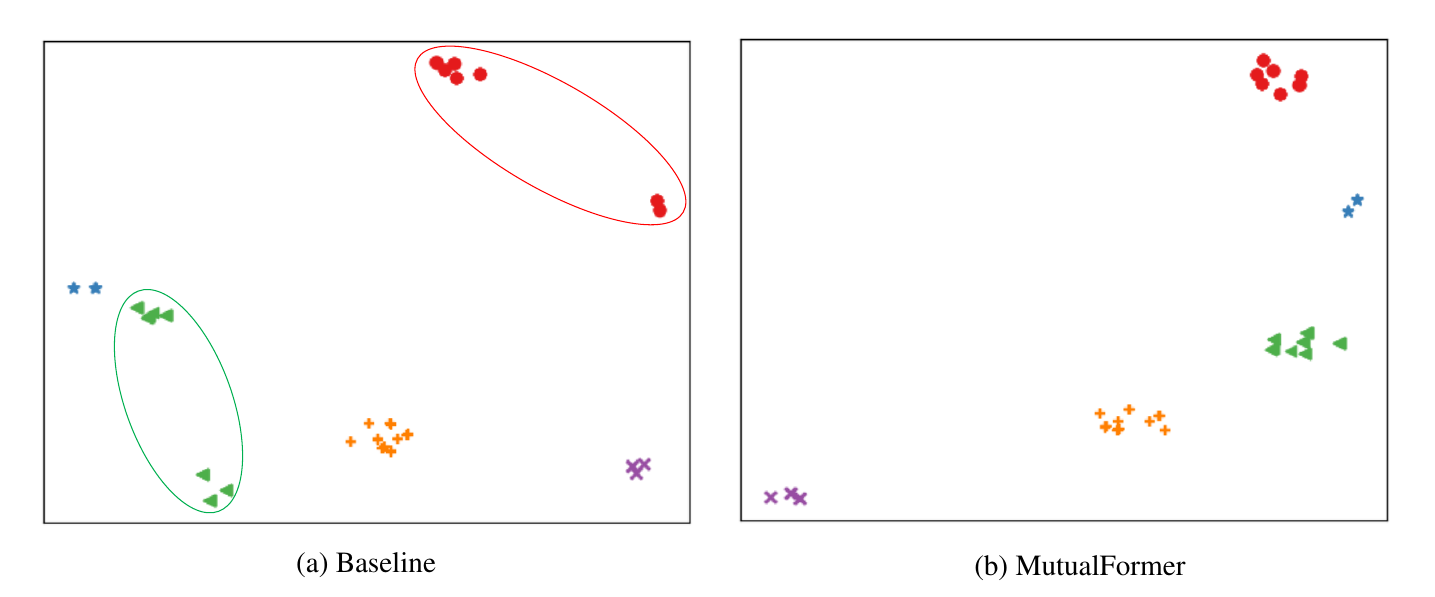}
\caption{2D t-SNE~\cite{van2008visualizing} visualization of the learned feature representations by MutualFormer and Baseline on RGBN300~\cite{li2020multi} dataset, respectively.
The different colors/shapes denote different identities/classes.
We can observe that the feature representation learned by the proposed MutualFormer is more discriminative than the
Baseline model.
}
\label{fig:tsne}
\end{figure*}

\subsection{Feature Visualization}
According to the aforementioned experimental analysis on multiple benchmark datasets, we can find that our proposed MutualFormer indeed helps the RGB-Depth SOD and RGB-NIR object ReID tasks.
To help readers have a more intuitive understanding of our model, for RGB-D SOD task, we give some visualizations of feature maps of our MutualFormer and compared baselines, including Add, Concatenate (Cat), Transformer, CrossFormer, and CrossFormer$_{CDA}$.
As shown in Fig. \ref{fig:feature_map}, we can find that the proposed MutualFormer activates more effective object areas and thus obtain better visual effects for RGB-D SOD task, which validate the effectiveness of our proposed MutualFormer.
In addition, we also give 2D visualizations of feature representation learned by our MutualFormer and Baseline model on RGBN300 dataset in RGB-NIR object ReID task.
As shown in Fig.~\ref{fig:tsne}, we show 5 identities/classes and use different colors/shapes to denote different classes.
We can observe that the feature representation learned by the proposed MutualFormer is more discriminative than the Baseline model.

In a word, these feature visualizations further demonstrate the effectiveness and superiority of the proposed MutualFormer for multi-modality data representation.

%
%

\section{Conclusion \& Future work}

This paper proposes a novel MutualFormer to conduct the interaction between tokens and modalities for multi-modality learning tasks.
Rather than using cross-attention, MutualFormer employs  Cross-Diffusion Attention (CDA) to conduct the information communication among different modalities which thus can
naturally avoid the
issue of domain gap.
Also, CDA is more flexible and efficient than CA.
In order to validate the effectiveness of our proposed MutualFormer, we take the RGB-D SOD and RGB-NIR object Re-ID tasks. 
Promising experiments demonstrate the effectiveness of our proposed CDA and MutualFormer.

In this paper, we mainly focus on multi-modality learning.
However, as we all know that,
cross-attention (CA) which is also used to capture the
dependence
between encoder and decoder
is the core module in regular Transformers and has been
widely used in various computer vision tasks.
In our future work, we will attempt to replace CA with our CDA to derive some more flexible and reliable Transformers for the various general computer vision problems.

\bibliography{reference}


\end{document}